# Image Processing Failure and Deep Learning Success in Lawn Measurement

**J. Wilkins, M. V. Nguyen, B. Rahmani**

Math and Computer Science Department, Fontbonne University

St. Louis, MO 63105, USA

**ABSTRACT**

Lawn area measurement is an application of image processing and deep learning. Researchers have been used hierarchical networks, segmented images and many other methods to measure lawn area. Methods effectiveness and accuracy varies. In this project Image processing and deep learning methods has been compared to find the best way to measure the lawn area. Three Image processing methods using OpenCV has been compared to Convolutional Neural network which is one of the most famous and effective deep learning methods. We used Keras and TensorFlow to estimate the lawn area. Convolutional Neural Network or shortly CNN shows very high accuracy (94-97%) for this purpose. In image processing methods, Thresholding with 80-87% accuracy and Edge detection are effective methods to measure the lawn area but Contouring with 26-31% accuracy does not calculate the lawn area successfully. We may conclude that deep learning methods especially CNN could be the best detective method comparing to image processing and other deep learning techniques.

# 1. Introduction

While there are definitely major business applications for lawn care companies, this project is a great experiment into an area of image processing that is rarely looked at: image regression. So, the project had to be started nearly from scratch, and creative ways were found to deal with the limited quality and quantity of the samples.

At the lawn care company that the author Jessie Wilkins worked at, the lawn area of houses from satellite images had to be measured. The process was rather simple: find the next address of the house, use online software to measure the area of the lawn of the house, record the measurements in an excel sheet, and repeat the steps. While the company sent out people to manually measure the lawn area, they wanted some rough estimate beforehand in areas. It was a very repetitive job, but it could only be done by a human. So, the lawn care company had to pay the person at least minimum wage, and the person had to spend many hours trying to measure hundreds of houses. Therefor the company was spending thousands of dollars and inordinate amount of time to measure houses. Eye-strain and fatigue are another issue that causes inconsistency and low accuracy. The lawn care company could save thousands of dollars and hundreds of human hours in measuring the lawns of houses accurately (at least within a margin of error).

Weiqi Zhou and his colleagues in 2008 tried to measure the lawn properties including lawn and house area remotely. They used hierarchical networks and classified segmented images for this purpose [1]. Alexander Schepelmann in his master thesis, measured the lawn area using color and visual texture classifier [2].

One of the methods that could be used by the lawn care company is artificial neural networks. Artificial neural networks have a long winding history in the computer science field. While the idea of a neural network in general has its basis in biology and did influence artificial neural networks, their mathematical conception dates back to 1943 in a paper of Warren S. McCullough and Walter Pitt [3]. In 1950s, several attempts were made to simulate a neural network by Clabaugh, et. al. In 1959, Stanford researchers Bernard Widrow and Marcian Hoff successfully implemented it in an algorithm [4]. They removed echoes and other data corruption from phone lines by predicting the next actual bit. In the 1980s, neural networks were revived as a topic of interest, and several advances to the artificial neural network architecture were made. After that productive decade, progress slowed down; however, due to even further developments in both artificial neural networks and increased processing power in computers, neural networks are now extremely popular and making large advances (Hardesty). Relevant to our query about measuring lawns from satellite images, convolutional neural networks are a major advancement in recent years for a variety of reasons.

Specifically, convolutional neural networks would work theoretically well because the lawn measurement problem is an image problem. The convolutional neural network pulls out the features such as grass texture, tree patterns, and other image features, the dense network layers analyze those features and determine the lawn area.

In this project, first we show the disability of image processing methods in calculating house and lawn area. Then we show that deep learning and convolutional neural network would work very well in this purpose. Data will be described in section 2. Section 3 discuss about theoretical background. Results and conclusions come after in section 4.

**2. Data Description**

Creating the neural network for the deep learning problem is only half of the battle. The other half is collecting all the necessary data. The specific dataset of satellite images of houses with their measured lawn area is a big challenge. While there were datasets that involved satellite imagery, they often involved much larger areas and usually implemented in classification problems. Despite this problem, there were imperfect but still effective ways of collecting the data needed for this project. The most immediate solution that was used and is still using is measuring the lawn area of houses on Google Maps and cropping these images out of the screen shots. While image quality was somewhat lower, it was the best dataset that could be found. We used online software Area Calculator (*Area Calculator Using Maps*) to measure each total lawn area in square meters. Then we used Krita, an image editing software ("Digital Painting. Creative Freedom") to crop the larger picture and make smaller individual pictures. The resulted images contain one of the measured areas with its surrounding lawn. In total, the author collected 65 pictures of houses and their lawns. This dataset was too small to actually use as training data. In order to remedy this problem, we used the ImageDataGenerator class mentioned below in order to duplicate each picture. In many deep learning image processing problems, artificial data is a legitimate way to reduce potential overfitting and increase performance. One way is to duplicate a picture but then slightly change it in some way in order to produce an essentially different picture; this can be done by image's rotating, inverting or flipping, distorting, changing the brightness of the image, and more. The ImageDataGenerator class has a method that can do all of those manipulations by just specifying in the constructor parameters of how you want to change each picture. The change is often random and within a given parameter range. This class was used iterate through each picture in the dataset and create an augmented duplicate of each image 50 times. Fig1 is a sample of duplicating images. 3000 images created with this method.

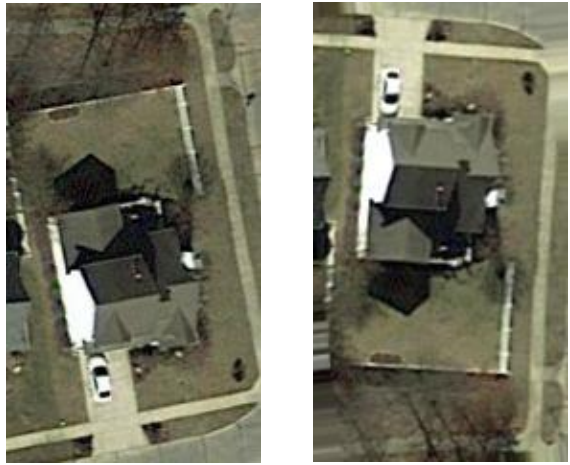

**Fig1   Left-Image before Data Augmentation and Right-Image after Data Augmentation**

We took a portion of created dataset as a test dataset. Test data for validation determines the overfitting.

## 3. Theoretical backgrounds

### 3.1. Convolutional Neural Network

Specifically, the convolutional neural network (CNN) helps to achieve a basis in deep learning projects. However, it is for classification rather than regression that is the main part of current problem. Most importantly, CNN introduces the Keras library which would be used to carry out the great majority of the process. Ultimately, the neural network did not reach the goal of 90% accuracy, but the accuracy was between 85 and 90%. Unfortunately, these did not necessarily explore overfitting in depth and still did not apply to the current work [5, 6]. So, we used Tensorflow that was more complex but too low-level to easily implement. There are many ways to implement convolutional neural networks; sometimes it can be done by defining and implementing complex functions with code from scratch [7] or by using Keras with Tensorflow as a backend [8]. Convolutional neural network is a 2D network meaning that it could take in 2-dimensional data. This is the kind of neural network used for image processing because pictures are represented as 2-dimensional through their height and width. Though the inputted arrays are 4 dimensions because it also includes the number of color channels and the number of pictures). It is 1-dimensional CNN for sentences and 3-dimensional CNN for voices. We start with three to four convolutional layers with two dense output layers at the bottom to process the information from the previous layers. After a few iterations, we used a different version of the convolutional layer in Keras that could supposedly get better performance called the separable convolutional layer. The separable convolutional layer or depthwise separable convolutional layer is an advanced version of the convolutional layer that can often increase the performance of deep learning image processing model [3].

Besides we implemented a new activation function in each layer called an ELU (instead of the previous RELU function) which again is often associated with being capable of producing better performance in neural networks [2]. Batch Normalization was also implemented because it can increase performance significantly, reduce the overfitting in the model, and deactivate a given layer in a certain percentage of the time [2]. We implemented regularizations at each layer which also are supposed to prevent overfitting [9]. Besides some of these more advanced implementations, we also added the usual convolutional layer complements such as pooling layers to help reduce the size of each picture at each layer [2]. The final model after extensive training on several models was a nine-layer convolutional neural network with 6 convolutional and 3 dense layers. This is probably the only neural network that specialized in regression analysis with the given satellite dataset. The basic procedure of this model is: an array containing all the picture arrays is passed into the first convolutional layer which pulls out certain low-level feature from the picture; the ELU activation function determines what values/features should be move to the next convolutional layer; a pooling layer reduces the size of the picture and passes from every two convolutional layers; and then, after the six convolutional layers, the features that are finally resulted as a flattened list are processed by the three Dense layers which determine the area that is correlated with the given feature [10]. By pulling out the features of the pictures through these convolutional layers and activations, the model was able to simplify the data and emphasize certain features. Fig2 shows an example.

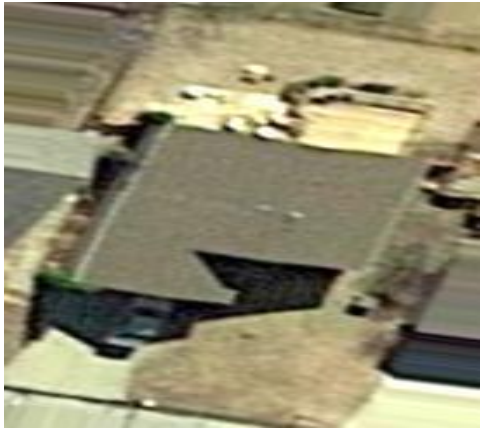

a)

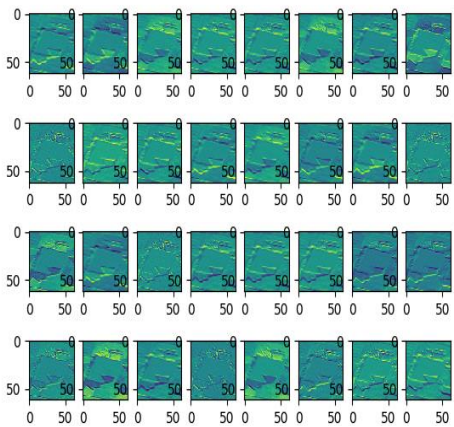

b)

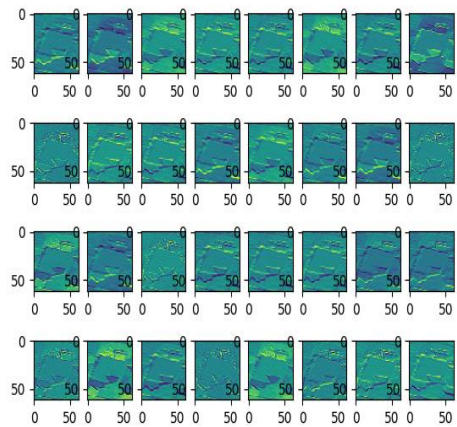

c)

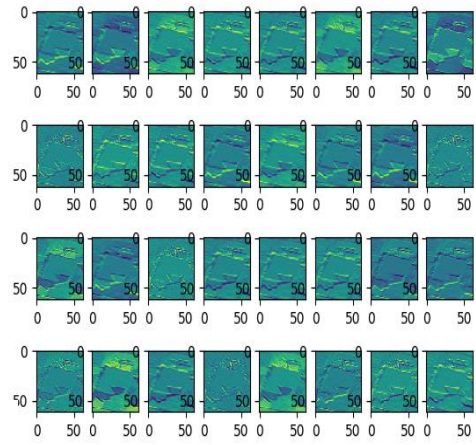

d)

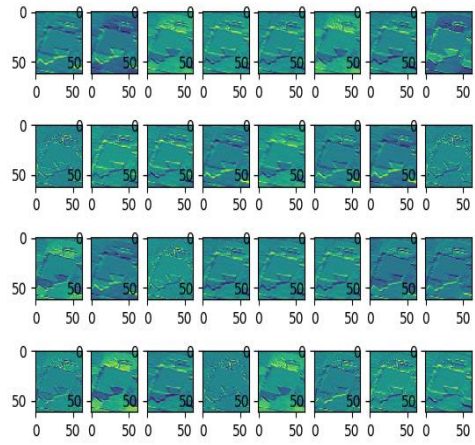

e)

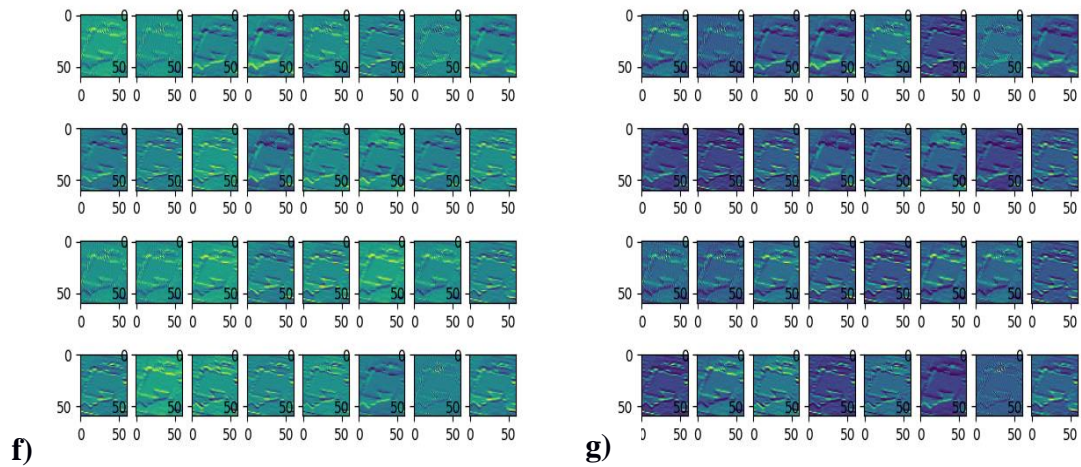

**Fig2  a) The original picture   b) Activation layer 1   c) Activation layer 2   d) Activation layer 3
e) Activation layer 4   f) Activation layer 5   g) Activation layer 6**

Fine-tuning of this neural network involved trial and error attempts of different number of neurons, layers, regularize parameters, and optimizers. Many of these parameters could be tested with a class from the Scikit-Learn library called GridSearchCV ("Sklearn.model_selection.GridSearchCV") [11]. GridSearchCV is a way of automating trial and error process as well as testing the true accuracy or performance of a neural network. One of the concepts in this library is cross-validation or CV. Essentially, CV takes the existing dataset and separates it into further sub-sets that can be used for validation. The GridSearch part of GridSearchCV test data by iterating through hyper-parameters. Hence, the GridSearchCV finds the best parameter too. While this method served its purpose well, it limited the dependability; specifically, in reducing overfitting. Because the validation partly relied on duplicated data from the same dataset, it does not show the best accuracy for reducing overfitting. Still, it was useful to see which parameters would over-fit the most and then try to reduce it by adjusting the dropout functions which calculates manually. Overall though, it led to the best performance as of now. Fig3 shows the final cropped model.

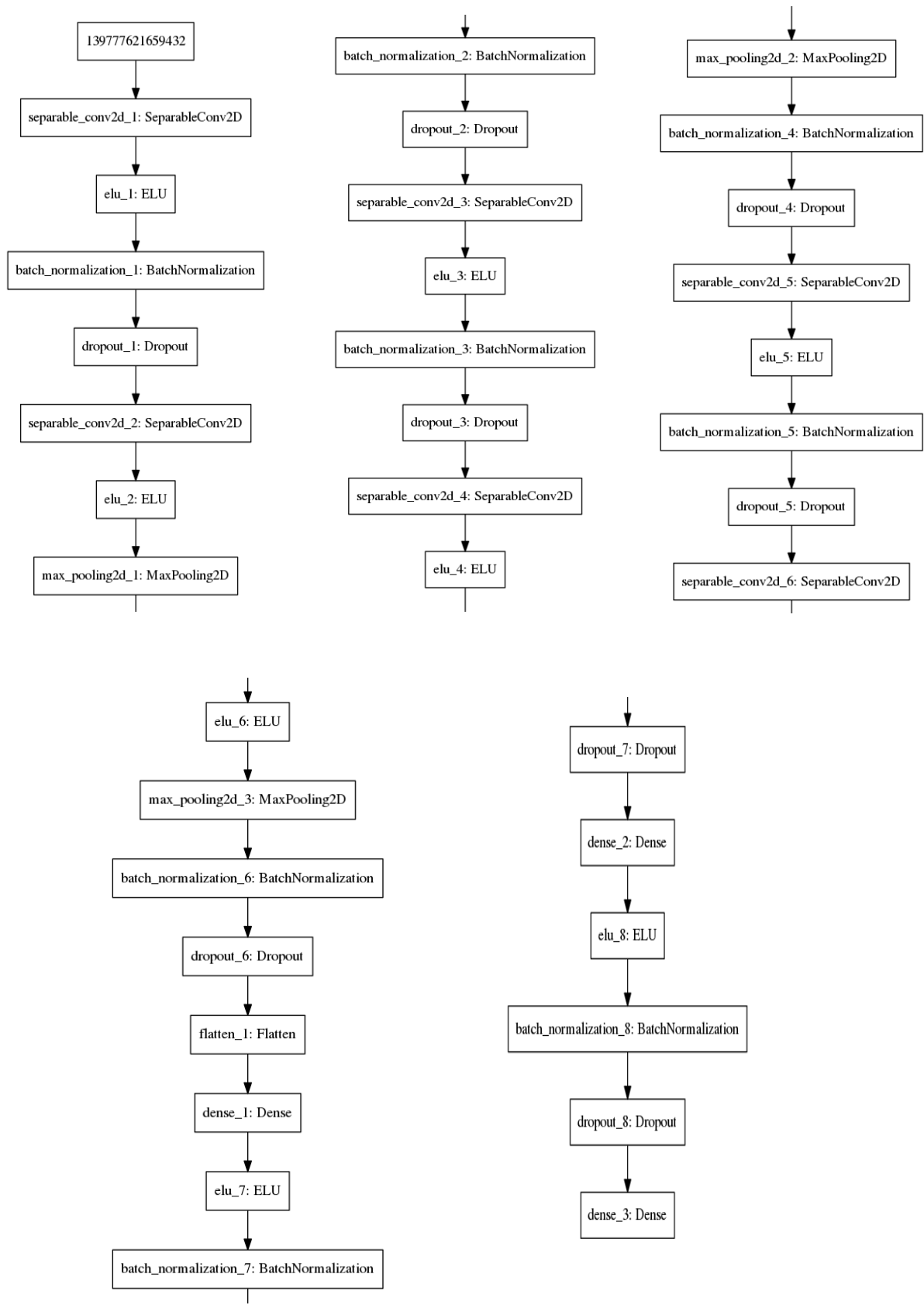

Fig3 The final cropped images of the main model used, the top of the model starts with the left most image ends at the bottom at the right-most image.

## 3.2. Thresholding and Augmentation

Though the model significantly fine-tuned, it was still below of 90% accuracy or even got to 85% especially on the test set. Therefore, we worked on another solution like OpenCV which is a very non-deep learning solution to the problem [12]. OpenCV is an image processing library in Python. It has several methods and classes used to manipulate images that can extract desired features as well as identifying them. OpenCV emphasize or remove unnecessary features from the pictures that let the neural network focuses on the essential features needed to measure the lawn area. Three different methods investigated: thresholding, finding contours, and edges [13]. Thresholding takes a certain range of pixel color density and turning it into the black/white image. The purpose of thresholding is to make the house in the middle of each picture totally in one color (white) and the rest of the lawn area to another color (black). Thresholding makes measuring of the lawn area simpler by removing extraneous features that had nothing to do with lawn area. Fig3 shows an Image before and after Thresholding and after both Thresholding and Augmentation.

## 3.3. Contouring and Augmentation

Unfortunately, after several tests and some fine-tuning, thresholding and augmentation did not work very well because they could not even replicate the current model's accuracy. The next method was finding the contours of the houses and their lawn area. Contour method finds the edges and curves of the houses (Fig4).

Despite adding extra information contouring also failed to exceed the performance of the original neural network.

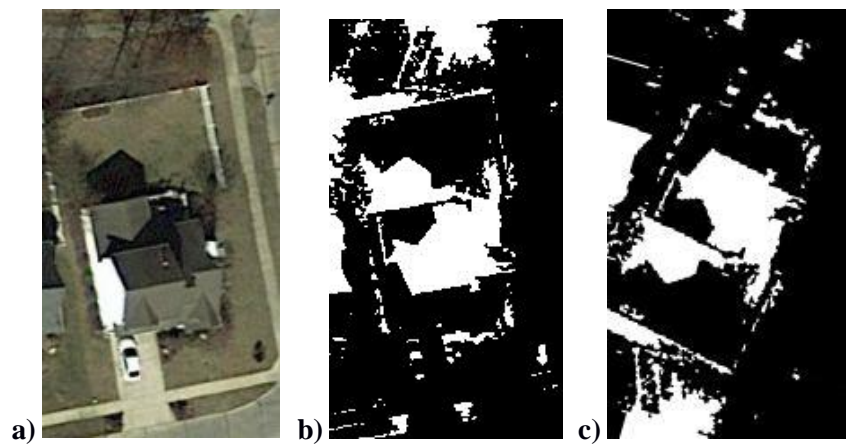

**Fig4   a) Image before Thresholding    b) Image after Thresholding
c) Image after Thresholding and Augmentation**

### 3.3. Contouring and Augmentation

Unfortunately, after several tests and some fine-tuning, thresholding and augmentation did not work very well because they could not even replicate the current model's accuracy. The next method was finding the contours of the houses and their lawn area. Contour method finds the edges and curves of the houses (Fig4).

Despite adding extra information contouring also failed to exceed the performance of the original neural network.

### 3.4. Canny Edge Algorithm

Finally, we used Canny Edge algorithm to find the edges of the house traced by the program that was able to simulate the desired result of the thresholding (Fig5).

Unfortunately, this algorithm also failed to meet expectations. Despite OpenCV being a powerful tool in many image processing applications, it was insufficient for this project's purpose.

### 4. Results

The final accuracy of convolutional neural network is not quite at 90%, but it definitely was able to measure lawn area within a given margin of error. Looking at the *Mean Squared Error (MSE)* of the prediction of the test data after the most recent training on 1,849 pictures, 1437 *MSE* is the given result which is around 38 square meters ($\sqrt{1437}$ = ~38 sq. m.) margin of error for each predicted lawn area. To get something like an accuracy percentage,

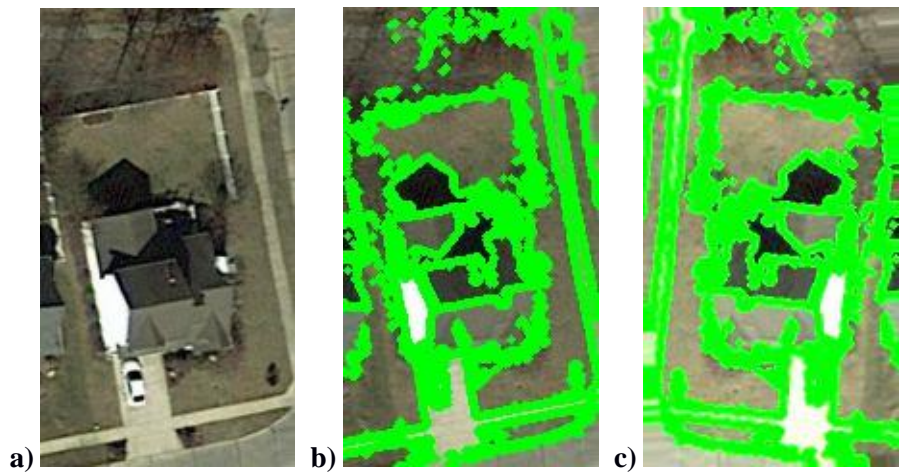

a)             b)             c)

**Fig5   a) Image before Contouring    b) Image after Contouring and Gaussian Blurring,
c) Image after Contouring and Augmentation**

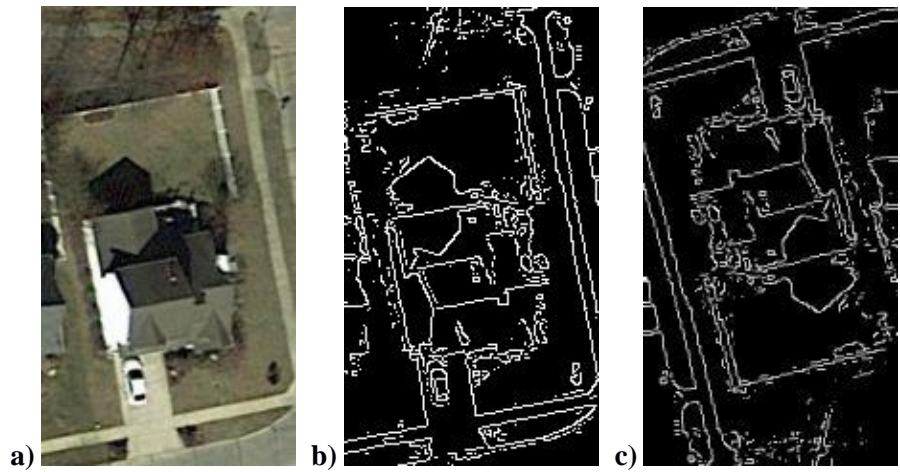

**Fig6  a) Image before Canny Edge Algorithm   b) Image after Canny Edge Algorithm
c) Image after Canny Edge Algorithm and Augmentation**

the ratio of the margin of error to the actual average value subtracted from 1 *(1-(error/average_val))*. The average measurement of the data that was trained on this model (pictures 1-45) is roughly 294 sq. m. The accuracy is on average ~87%  (1-(38/294) = ~87%). The median, which is approximately 276 sq. m. with the accuracy: ~86%  (1-(38/276) = ~86%). The test result unfortunately is the result of some overfitting that has been occurring, but it is not that radically different from the training results. The mean squared error of the training vs the testing is only off by a less than 100 *MSE* even though there is a greater difference between training and validation *MSE*. However, there is some evidence of volatility in this model; for example, in another test, the *MSE* was 2366 meaning an average error of ~49 Squared meters, ~83% average accuracy, and ~83% median accuracy. While this would not be acceptable to any lawn care business yet, this is still a sign of hope that this neural network can at least predict the value within a reasonable margin of error.

| Model Used | Highest Accuracy (1- (error/Average of Original Data)) | Model Results (Average Predicted Lawn Area) | Average Lawn Area of Used Data |
|---|---|---|---|
| CNN Training | - | -* | ~298.97 |
| CNN Validation | ~94% | ~280.12 m$^2$ | ~298.14 m$^2$ |
| CNN Testing | ~97% | ~262.65 m$^2$ | ~254.17 m$^2$ |
| Threshold Model Training | - | -* | ~283.55 |
| Threshold Model Validation | ~80% | ~228.87 | ~287.21 |
| Threshold Model Testing | ~87% | ~221.22 | ~254.17 |
| Contour Model Training | - | -* | ~297.66 |
| Contour Model Validation | ~26% | ~634.49 | ~280.35 |
| Contour Model-Testing | ~31% | ~587.50 | ~254.17 |
| Edges Model Training | - | -* | ~297.89 |
| Edges Model Validation | ~85% | ~237.80 | ~280.53 |
| Edges Model Testing | ~57% | ~145.16 | ~254.17 |

**Table 1 Mean Squared Error (MSE) of training, validation, and testing datasets**

* The average predicted result was not shown for the training data since the MSE recorded during training does not correspond with the MSE and the predicted lawn area observed through Keras' predict function for the training data.

As mentioned before, there are 65 pictures accumulated. So, there is approximately 3000 duplicated pictures in total. However, the final result used only the 45 original pictures which were separated into training, validation, and test data. In this case, there were really 1,849 training samples, 150 validation samples, and 250 test samples when this neural network achieved the above accuracy. The results obtained by image processing methods have almost half accuracy of deep learning methods.